\documentclass{article}
\usepackage{spconf,amsmath,epsfig}
\usepackage[pagebackref=true,breaklinks=true,letterpaper=true,colorlinks,bookmarks=false]{hyperref}
\usepackage{graphicx}
\usepackage{enumitem}
\setlist{nosep}
\title{Sen2Fire: A Challenging Benchmark Dataset for Wildfire Detection using Sentinel Data}
\name{Yonghao Xu$^{1*}$, Amanda Berg$^{1,2}$, and Leif Haglund$^{1,2}$\thanks{$^{*}$This work was supported in part by the Excellence Center at Linköping-Lund in Information Technology (ELLIIT) Researcher Funding, the Zenith Research Program, and the Vinnova Advanced and Innovative Digitalization Project under Grant 2023-01904.}}
\address{$^1$Computer Vision Laboratory, Linköping University, Sweden\\$^2$Maxar Technologies, Sweden}

\begin{document}
%\ninept
%
\maketitle
\begin{abstract}
Utilizing satellite imagery for wildfire detection presents substantial potential for practical applications. To advance the development of machine learning algorithms in this domain, our study introduces the \textit{Sen2Fire} dataset--a challenging satellite remote sensing dataset tailored for wildfire detection. This dataset is curated from Sentinel-2 multi-spectral data and Sentinel-5P aerosol product, comprising a total of 2466 image patches. Each patch has a size of 512$\times$512 pixels with 13 bands. Given the distinctive sensitivities of various wavebands to wildfire responses, our research focuses on optimizing wildfire detection by evaluating different wavebands and employing a combination of spectral indices, such as normalized burn ratio (NBR) and normalized difference vegetation index (NDVI). The results suggest that, in contrast to using all bands for wildfire detection, selecting specific band combinations yields superior performance. Additionally, our study underscores the positive impact of integrating Sentinel-5 aerosol data for wildfire detection. The code and dataset are available online (https://zenodo.org/records/10881058).
\end{abstract}
\section{Introduction}
\label{sec:intro}

Wildfires are unplanned and uncontrolled fires in an area of combustible vegetation \cite{team2017canadian}. As a global natural disaster, wildfires present a substantial threat to both the ecological environment and the social economy \cite{jolly2015climate,kondylatos2022wildfire}. For example, it is reported that California’s 2018 wildfires alone cost the US economy 148.5 billion dollars, which accounts for 0.7\% of the country’s GDP \cite{wang2021economic}. Therefore, rapid and accurate wildfire detection emerges as a pivotal research focus.% within the realm of environmental science.

Recently, the integration of satellite imagery in wildfire monitoring, particularly when coupled with deep learning algorithms, has opened up new avenues for enhancing monitoring precision and efficiency \cite{huot2022next,eddin2023location}. Nevertheless, the existing research on remote sensing-based wildfire detection encounters several challenges \cite{rashkovetsky2021wildfire}. These include the lack of large-scale benchmark datasets, the distinct sensitivities of different wavebands to wildfires, and the transferability of models across diverse geographical locations.

\begin{figure}[t]
\centering
\includegraphics[width=\linewidth]{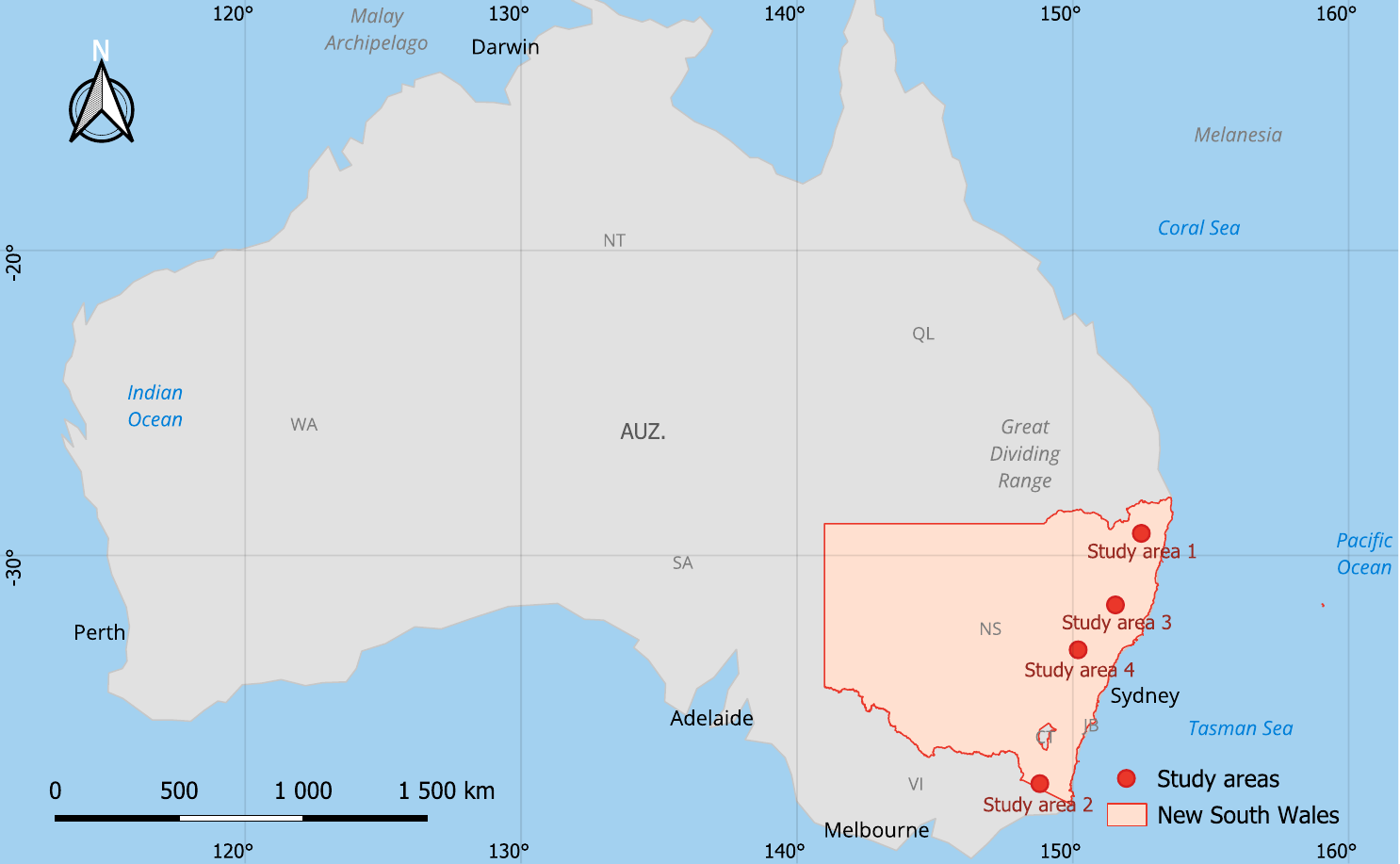}
\caption{Study areas of the \textit{Sen2Fire} dataset: Four bushfires happened in the 2019--2020 Australian bushfire season.}
\label{fig:study_area}
\end{figure}

To address the aforementioned challenges and advance the development of machine learning algorithms in this domain, we introduce the \textit{Sen2Fire} dataset–-a satellite remote sensing dataset tailored for wildfire detection. \textit{Sen2Fire} is curated from both Sentinel-2 multi-spectral data and Sentinel-5P aerosol product. Considering that the occurrence of wildfires is often accompanied by the production of large amounts of aerosols \cite{solomon2023chlorine}, we hope that the introduced sentinel-5P data in \textit{Sen2Fire} can provide data support for research on wildfire detection based on satellite aerosol products. Figure~\ref{fig:study_area} demonstrates the detailed geographic locations of study areas of this dataset (see Section~\ref{sec:areas} for details), which contains four bushfires in New South Wales, Australia, happened in the 2019--2020 Australian bushfire season.

\begin{figure*}[t]
\centering
\includegraphics[width=.95\linewidth]{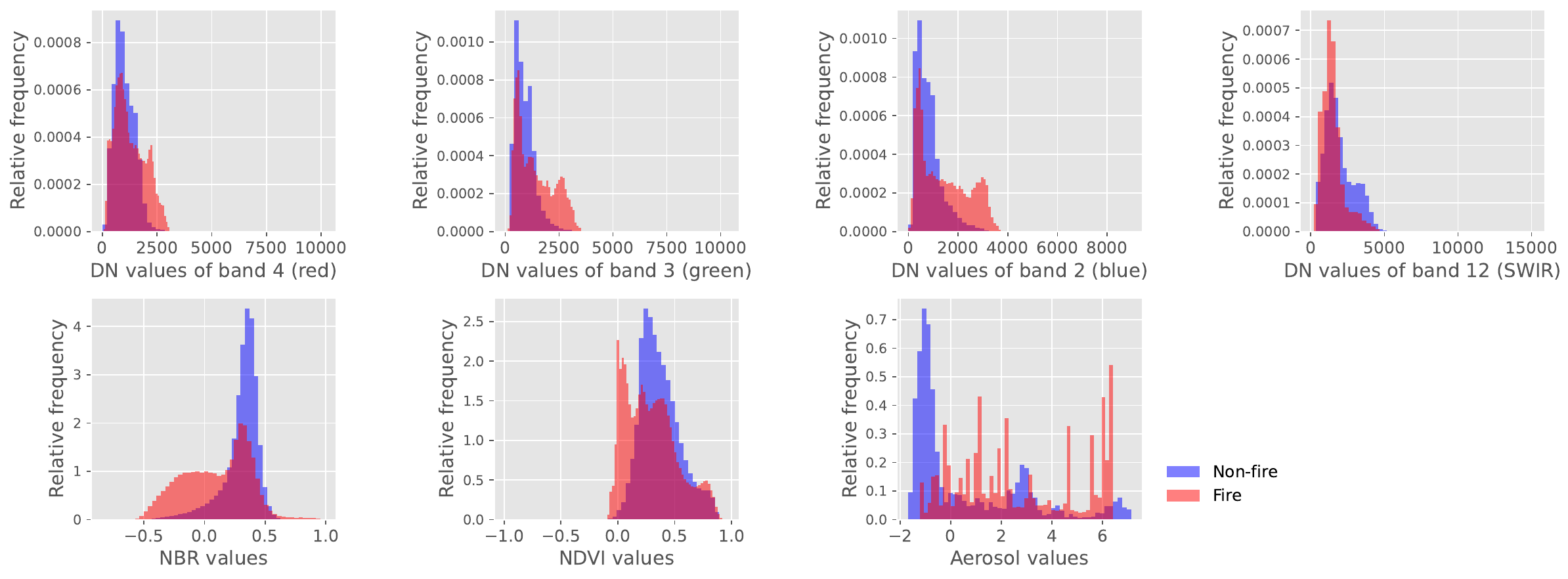}
\caption{The relative frequency distribution of the digital number (DN), index, or aerosol values for fire and non-fire samples in the training set. There exist high overlaps (denoted in \textcolor[rgb]{.7,0,0}{dark red}) between the distributions of fire and non-fire samples.}
\label{fig:hist}
\end{figure*}

Our contributions are summarized as follows:
\begin{itemize} 
    \item We introduce a challenging satellite remote sensing dataset for wildfire detection by fusing Sentinel-2 multi-spectral data and Sentinel-5P aerosol product.
    \item We studied the impact of different waveband and spectral indices combinations on wildfire detection performance. Compared to the traditional strategy of fusing all multi-spectral bands as model input, our experimental results show that selecting specific band combinations yields superior performance.
    \item Our experimental results demonstrate the positive impact of integrating satellite aerosol products on wildfire detection, which brings about a new potential for remote sensing-based wildfire detection research.
\end{itemize}

\section{Dataset Description}
\subsection{Study Areas}
\label{sec:areas}
As shown in Figure~\ref{fig:study_area}, the study areas of \textit{Sen2Fire} are located in four regions of the state of New South Wales in Australia. These wildfires happened in the 2019–-2020 Australian bushfire season, which is one of the most intense and catastrophic fire seasons on record in Australia \cite{deb2020causes}. Over 55,000 square kilometers, nearly 7\% of the state, were consumed by bush and grass fires. It led to the destruction of 2,448 homes, and 26 lives were lost during the fire season \cite{NSWFireSeason2020}.
The selected study area covers a total of more than 30,000 square kilometers from southern to northeastern New South Wales and thereby, can provide a strong data benchmark for evaluating the geographical transferability of wildfire detection models.

\subsection{Satellite Data}
\label{sec:satellite_data}
We exploit freely available satellite data or derived products to form the basis of the \textit{Sen2Fire} dataset. Specifically, we collect Sentinel-2 multi-spectral imagery and Sentinel-5P aerosol products to serve as the input variables. The ground-truth labels are collected based on the MOD14A1 V6.1 global daily fire product.

\textit{Sentinel-2} is an Earth observation mission developed by the European Space Agency (ESA) that systematically acquires optical imagery at high spatial resolution (10 to 60 meters) over land and coastal waters \cite{drusch2012sentinel}. It provides multi-spectral data with 13 bands in the visible, near infrared (NIR), and short-wave infrared (SWIR) parts of the spectrum. For the \textit{Sen2Fire} dataset, we provide the full multi-spectral image cubes (12 bands) as extracted from the original Sentinel-2 Level-2A data (note that the cirrus band is used for atmospheric correction only, thus abandoned in Level-2A data).

\textit{Sentinel-5 Precursor (Sentinel-5P)} is an Earth observation satellite developed by ESA for air pollution monitoring \cite{veefkind2012tropomi}. It provides multi-spectral data with 7 bands in ultraviolet (UV), visible, NIR, and SWIR parts of the spectrum. Based on the wavelength-dependent changes in Rayleigh scattering in the UV spectral range, Sentinel-5P data can be used to measure the global aerosols. For the \textit{Sen2Fire} dataset, we use the Sentinel-5P UV aerosol index product as an additional input layer with a spatial resolution of around 1 kilometer.

\textit{MOD14A1 V6.1} is a global daily fire product developed by the National Aeronautics and Space Administration (NASA) \cite{justice2002modis}. It provides daily fire masks derived from the Moderate Resolution Imaging Spectroradiometer (MODIS) 4- and 11-micrometer radiances. For the \textit{Sen2Fire} dataset, we use the MOD14A1 V6.1 product as the ground-truth label of the wildfires with a spatial resolution of 1 kilometer.

\begin{figure}[t]
\centering
\includegraphics[width=.9\linewidth]{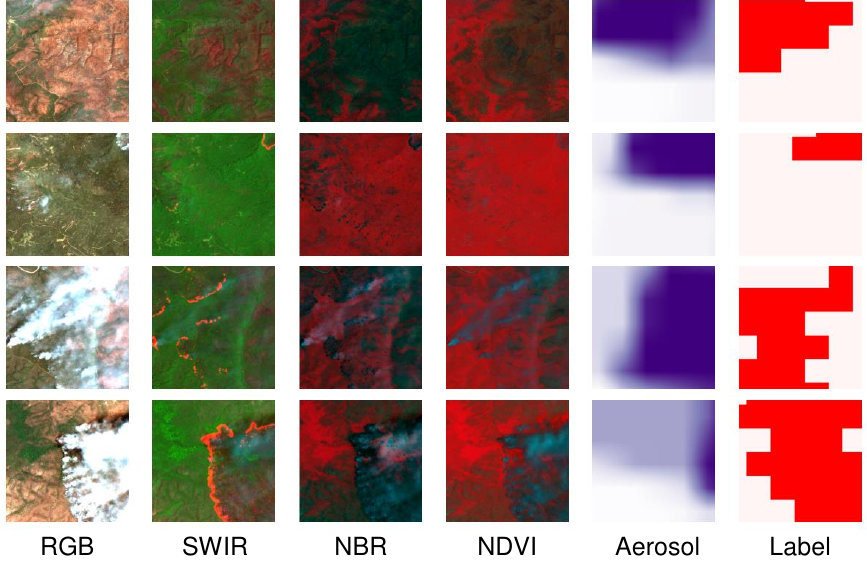}
\caption{Visualization for patches in the training set. RGB, SWIR, NBR, and NDVI denote different color composites.% (see Section~\ref{sec:input} for details).
}
\label{fig:demo}
\end{figure}

\begin{table}
\centering
\caption{Band information in the \textit{Sen2Fire} dataset.}
\label{tab:band}
\vspace{.3em}
\resizebox{\linewidth}{!}{%
\begin{tabular}{lcc}
\hline
Sen2Fire   bands          & Central wavelength ($\mu m$) & Original resolution ($m$)  \\
\hline
B1 -- Coastal aerosol     & 0.443                        & 60                       \\
B2 -- Blue                & 0.490                        & 10                       \\
B3 -- Green               & 0.560                        & 10                       \\
B4 -- Red                 & 0.665                        & 10                       \\
B5 -- Vegetation red edge & 0.705                        & 20                       \\
B6 -- Vegetation red edge & 0.740                        & 20                       \\
B7 -- Vegetation red edge & 0.783                        & 20                       \\
B8 -- NIR                 & 0.842                        & 10                       \\
B9 -- Vegetation red edge & 0.865                        & 20                       \\
B10 -- Water vapour       & 0.945                        & 60                       \\
B11 -- SWIR               & 1.610                        & 20                       \\
B12 -- SWIR               & 2.190                        & 20                       \\
B13 -- Aerosol index      & /                            & 1113                     \\
\hline
\end{tabular}%
}
\end{table}

\subsection{Data Curation}
For every study area depicted in Figure~\ref{fig:study_area}, we acquire satellite data, as detailed in Section~\ref{sec:satellite_data}, on the same day. Subsequently, all collected data is resampled to achieve a consistent 10-meter spatial resolution. The training set is composed of data from study areas 1 and 2, the validation set is derived from study area 3, and study area 4 is designated as the test set. It's important to note that there is no overlap between different study areas to prevent potential label leakage issues during the model training process. Following this, each study area is partitioned into 512$\times$512 patches, with an overlap of 128 pixels between adjacent patches. After this tiling process, the training, validation, and test sets consist of 1458, 504, and 504 patches, respectively. Each patch within these sets includes 13 channels, comprising 12 multispectral channels from Sentinel-2 and 1 aerosol index product channel from Sentinel-5, as shown in Table~\ref{tab:band}. Additionally, each patch is associated with a wildfire label, where pixel values of 0 and 1 signify non-fire and fire, respectively.

\begin{figure}[t]
\centering
\includegraphics[width=\linewidth]{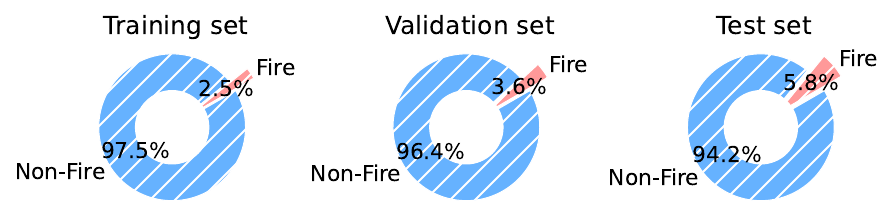}
\caption{The ratio of the fire samples to the non-fire samples.}
\label{fig:ratio}
\end{figure}

Figure~\ref{fig:hist} illustrates the relative frequency distribution of the DN, spectral indices (e.g., NBR, as detailed in Section~\ref{sec:index}), and aerosol values for both fire and non-fire samples within the training set. Notably, there exist substantial overlaps between the distributions of fire and non-fire samples, which poses a significant challenge in distinguishing between these two categories.
In Figure~\ref{fig:demo}, we provide visualizations of randomly selected patches from the training set, accompanied by their wildfire labels. In addition to the standard RGB composite, we also generate false color composites for visualization based on the SWIR band or spectral indices (see Section~\ref{sec:input} for details).
Figure~\ref{fig:ratio} further displays the ratio of fire category samples to non-fire samples across the training, validation, and test sets. A noteworthy observation is the pronounced class imbalance in all subsets of \textit{Sen2Fire}.

\section{Methodology}
Considering the distinct sensitivities of different wavebands to wildfires, this study aims to explore the impact of different waveband combinations on wildfire detection performance. Some well-defined spectral indices that are related to burned or vegetation areas are also considered.
\subsection{Spectral Indices}
\label{sec:index}
We take the normalized burn ratio (NBR) and the normalized difference vegetation index (NDVI) as the main spectral indices to assist wildfire detection.

NBR is commonly used to highlight burned areas and provide a measure of burn severity, which is defined as:
\begin{equation}
NBR = \frac{NIR - SWIR}{NIR + SWIR}~,
\label{eq:nbr}
\end{equation}
where $NIR$ and $SWIR$ stand for the spectral reflectance measurements acquired in the near infrared and short-wave infrared regions, respectively. NBR values close to 1 usually indicate healthy vegetation and values near or below 0 often indicate burned or unproductive vegetation \cite{wulder2009characterizing}.

NDVI is a widely used metric for quantifying the health and density of vegetation using spectrometric data, which is defined as:
\begin{equation}
NDVI = \frac{NIR - Red}{NIR + Red}~,
\label{eq:ndvi}
\end{equation}
where $Red$ denotes the spectral reflectance measurements acquired in the red (visible) region. High NDVI values usually indicate dense vegetation with high chlorophyll concentration in leaves, while low NDVI values suggest sparse or stressed vegetation and non-vegetated surfaces. Thus, an abrupt decline in NDVI may indicate burnt vegetation \cite{kasischke1993monitoring}.

\begin{figure*}[t]

\centering
\includegraphics[width=.85\linewidth]{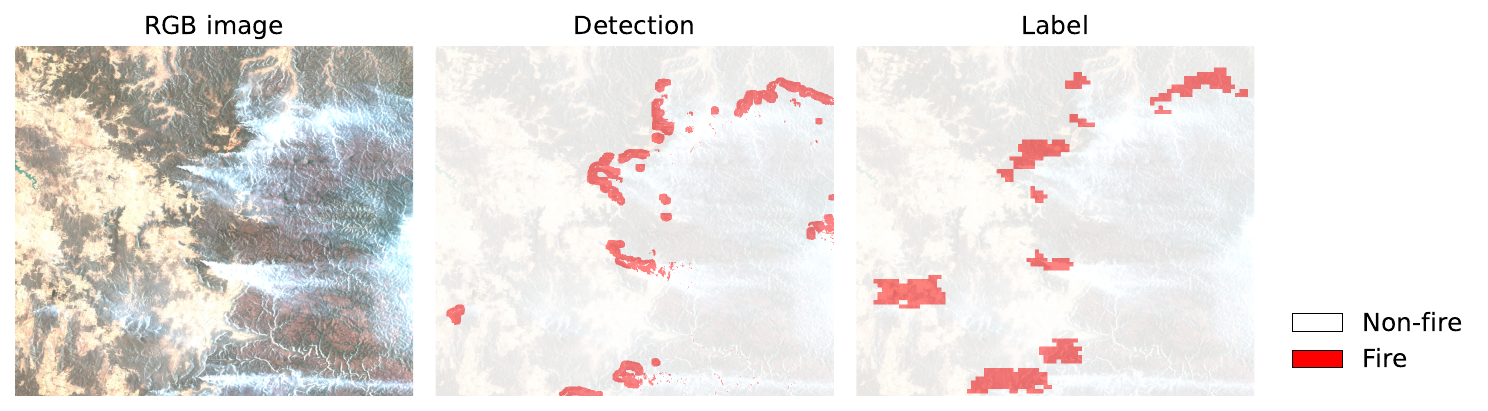}
\caption{Wildfire detection result on the test set. The input patches are concatenated to reconstruct the complete image tile.}
\label{fig:map}
\end{figure*}

\subsection{Input Strategies}
\label{sec:input}
Satellite multi-spectral data often have information redundancy. Applying suitable band selection strategies to select the most informative spectral bands may help reduce information redundancy and avoid the negative effect of irrelevant channels, as observed in a previous landslide detection research \cite{ghorbanzadeh2022outcome}. In this study, we explore the impact of different waveband and spectral indices combinations on wildfire detection performance based on the following 6 input strategies:
\\ \textit{RGB composite:} \texttt{B4,B3,B2}.
\\ \textit{SWIR composite:} \texttt{B12,B8,B4}.
\\ \textit{NBR composite:} \texttt{NBR,B4,B3}.
\\ \textit{NDVI composite:} \texttt{NDVI,B4,B3}.
\\ \textit{RGB+SWIR+NBR+NDVI:} \texttt{B4,B3,B2,B12,NBR,NDVI}.
\\ \textit{Vanilla input:} \texttt{B1,B2,B3,...,B10,B11,B12}.

The \textit{SWIR composite} is adopted considering its high sensitivity to high-temperature heat sources. Note that \textit{Vanilla input} corresponds to the traditional strategy of fusing all 12 multi-spectral bands. For each strategy, we also compare its performance w/ and w/o the integration of Sentinel-5P aerosol product, resulting in 12 input combinations in total.
\section{Experiments}

\begin{table}[]
\centering
\caption{Quantitative results (\%) obtained by U-Net with different input strategies for wildfire detection on the test set.}
\label{tab:result}
\resizebox{.9\linewidth}{!}{%
\begin{tabular}{cccc}
\hline
Input strategies  & Precision & Recall & F1 score   \\
\hline
RGB composite     & 11.8      & 18.3   & 14.4 (*) \\
+aerosol          & 14.7      & 21.3   & 17.4\textcolor[rgb]{0,0.6,0}{$_{\uparrow 3.0}$} \\
\hline
SWIR composite    & 43.9      & 20.5   & 27.9 (*) \\
+aerosol          & 39.7      & 21.8   & 28.1\textcolor[rgb]{0,0.6,0}{$_{\uparrow 0.2}$} \\
\hline
NBR composite     & 26.0      & 24.1   & 25.1 (*) \\
+aerosol          & 20.6      & 23.9   & 22.1\textcolor[rgb]{1,0,0}{$_{\downarrow 3.0}$} \\
\hline
NDVI composite    & 13.4      & 13.0   & 13.2 (*) \\
+aerosol          & 11.4      & 23.1   & 15.2\textcolor[rgb]{0,0.6,0}{$_{\uparrow 2.0}$} \\
\hline
RGB+SWIR+NBR+NDVI & 38.6      & 17.1   & 23.7 (*) \\
+aerosol          & 35.5      & 19.1   & 24.8\textcolor[rgb]{0,0.6,0}{$_{\uparrow 1.1}$} \\
\hline
Vanilla input     & 22.4      & 29.5   & 25.5 (*) \\
+aerosol          & 37.4      & 20.1   & 26.1\textcolor[rgb]{0,0.6,0}{$_{\uparrow 0.6}$} \\
\hline
\end{tabular}%
}
\\\vspace{1pt}
\raggedright \scriptsize \noindent \textcolor[rgb]{0,0.5,0}{Green}/\textcolor[rgb]{1,0,0}{red}: F1 score gain/loss by the integration of aerosol product with respect to the corresponding input strategy baseline (*).
\end{table}

\subsection{Implementation Details}
We utilized the U-Net \cite{ronneberger2015u} architecture as our foundation model to evaluate wildfire performance across different input strategies. The model was trained using the Adam optimizer, employing a learning rate of $1e-4$ and a batch size of 8 over 5000 iterations. All experiments were executed on a cluster using a single NVIDIA A100 GPU. During the training phase, the validation set was employed to select the optimal model, and the accuracy of the test set was recorded as the final result for each input strategy. %The assessment of wildfire detection performance on the test set involved three standard accuracy metrics: a) precision, b) recall, and c) F1-score. All metrics are calculated based on the predictions on the fire category.

\subsection{Experimental Results}
Table~\ref{tab:result} reports the quantitative results obtained by U-Net with different input strategies for wildfire detection on the test set. Compared to the traditional strategy of fusing all multi-spectral bands as model input, we find that selecting specific band combinations yields superior performance. While the F1 score obtained by vanilla input is 25.5\%, the SWIR composite can yield an F1 score of 27.9\%, which outperforms the former one by over 2\%. This phenomenon reveals the importance of designing appropriate input strategies for multi-spectral data. Besides, we find that the integration of aerosol products can further contribute to the performance of most input strategies in the experiment, which brings about a new potential avenue for satellite remote sensing-based wildfire detection research. Figure~\ref{fig:map} visualizes the qualitative result on the test set.

\section{Conclusions and Discussions}
In this study, we introduce the \textit{Sen2Fire} dataset, a curated and challenging collection derived from Sentinel-2 multi-spectral data and Sentinel-5P aerosol product for wildfire detection. Our experimental results suggest that, in contrast to the conventional approach of incorporating all multi-spectral bands for wildfire detection, selecting specific band combinations yields superior performance. Furthermore, our study underscores the positive impact of integrating Sentinel-5 aerosol data on wildfire detection, which provides a promising avenue for future remote sensing-based research in this domain. 

While this work makes a preliminary analysis of the impact of various waveband/spectral indices combinations on wildfire detection, the development of more effective strategies deserves further study. We will explore it in the future. 
%Given that \textit{Sen2Fire} ground-truth labels are obtained from MODIS fire products, potential errors may arise in model training due to spatial resolution variations. Consequently, investigating strategies, such as weakly supervised learning, to mitigate inaccuracies stemming from label discrepancies may further enhance wildfire detection performance. We will explore it in our future work.

\bibliographystyle{IEEEbib}
\bibliography{refs}

\end{document}